\documentclass[journal]{IEEEtran}
\usepackage{amsmath,amssymb,amsfonts}
\usepackage{algpseudocode,algorithm}
\usepackage{subfigure}
\usepackage{graphicx}
\usepackage{cite}

\begin{document}
%
\title{Latent Representation in Human-Robot Interaction with Explicit Consideration of Periodic Dynamics}

%

\author{Taisuke~Kobayashi,~\IEEEmembership{Member,~IEEE},
        Shingo~Murata,~\IEEEmembership{Member,~IEEE},
        and~Tetsunari~Inamura,~\IEEEmembership{Member,~IEEE}
\thanks{T. Kobayashi is with the Division of Information Science, Nara Institute of Science and Technology, 8916-5 Takayama-cho, Ikoma, Nara 630-0192, Japan
e-mail: (kobayashi@is.naist.jp).}
\thanks{S. Murata is with the Department of Electronics and Electrical Engineering, Faculty of Science and Technology, Keio University, Japan
e-mail: (murata@elec.keio.ac.jp).}
\thanks{T. Inamura is with National Institute of Informatics, Japan; and with The Graduate University for Advanced Studies (SOKENDAI), Japan
e-mail: (inamura@nii.ac.jp).}
}

\markboth{Preprint submitted to IEEE Transactions on Human-Machine Systems}%
{Kobayashi \MakeLowercase{\textit{et al.}}: Latent Space Analysis on Periodic Physical Human-Robot Interaction}
%



\maketitle

\begin{abstract}

This paper presents a new data-driven framework for analyzing periodic physical human-robot interaction (pHRI) in latent state space.
To elaborate human understanding and/or robot control during pHRI, the model representing pHRI is critical.
Recent developments of deep learning technologies would enable us to learn such a model from a dataset collected from the actual pHRI.
Our framework is developed based on variational recurrent neural network (VRNN), which can inherently handle time-series data like one pHRI generates.
This paper modifies VRNN in order to include the latent dynamics from robot to human explicitly.
In addition, to analyze periodic motions like walking, we integrate a new recurrent network based on reservoir computing (RC), which has random and fixed connections between numerous neurons, with VRNN.
By augmenting RC into complex domain, periodic behavior can be represented as the phase rotation in complex domain without decaying the amplitude.
For verification of the proposed framework, a rope-rotation/swinging experiment was analyzed.
The proposed framework, trained on the dataset collected from the experiment, achieved the latent state space where the differences in periodic motions can be distinguished.
Such a well-distinguished space yielded the best prediction accuracy of the human observations and the robot actions.

\end{abstract}


%
\IEEEpeerreviewmaketitle

\section{Introduction}

\IEEEPARstart{A}{s} the birthrate declines and the population ages, robots to supplement human labor are getting highly desired.
Such robots are required to perform tasks in real world with more uncertainty especially with physical human-robot interaction (so-called pHRI)~\cite{beckerle2017human}, such as in the welfare and service industries, instead of tasks in predetermined environments like factory automation.
A growing trend toward the development of practical pHRI-capable robots makes the studies on pHRI more active: for example, since simulation is essential for the development of robots, a simulation platform that can handle pHRI has been developed and served~\cite{inamura2021sigverse}.
Alternatively, human understanding through pHRI by robot collecting data about interaction with human is an open and challenging issue~\cite{kidd2008robots,liu2017understanding}.

For handling physical interaction, robot is often controlled with a virtual impedance model~\cite{khoramshahi2019dynamical,ferraguti2019variable,itadera2021towards}.
It enables robot to follow to human easily, but there are strong limitations on the contact points and the actions that can be generated.
Even changing the impedance parameter adaptively like the above literature, these limitations are imposed by the model itself and cannot be solved essentially.
Although the work on robots with tactile sensors all over the body has been studied~\cite{kobayashi2019multi,leboutet2019tactile}, their controller is still built based on a pre-assumed model and does not incorporate any particular human characteristics.

In contrast, recent remarkable developments of deep learning technologies~\cite{lecun2015deep} would shed light on modeling of pHRI from a dataset collected from the actual interactions.
Methods for extracting the latent state space hidden in high-dimensional observation data have been established~\cite{kingma2014auto,higgins2016beta}.
They have been applied to, for example, control applications based on learning of dynamics model~\cite{hafner2019learning,li2021planning} and motion classification in the extracted space~\cite{itadera2021towards,chen2016dynamic}.

In this study, we propose a new data-driven framework for modeling pHRI based on variational recurrent neural network (VRNN)~\cite{chung2015recurrent}, which is one of the techniques for extracting latent state space and can handle time-series data by integrated with recurrent neural networks (RNNs)~\cite{hochreiter1997long,murata2013learning}.
VRNN is first extended to be suitable for pHRI by explicitly considering the dynamics from robot to human in the extracted latent state space, while simplifying its computational graph for stable learning.
This extension derives a new variational lower bound.

Furthermore, we focus on the fact that most of the pHRI tasks show periodicity:
such as, polishing~\cite{khoramshahi2019dynamical}, brushing~\cite{ferraguti2019variable}, walking~\cite{itadera2021towards}, dancing~\cite{kobayashi2019multi}, and so on.
For simplifying the computational graph as mentioned above, reservoir computing (RC)~\cite{jaeger2004harnessing,lukovsevivcius2009reservoir,gallicchio2017deep,kobayashi2018practical}, a type of RNNs consisting of numerous neurons connected each other by random and fixed parameters, is first employed.
For inherently representing periodicity, although its parameters and internal states of neurons are originally given in real domain, we augment them to complex domain~\cite{goh2007augmented,hu2012global}, so-called complex-valued RC or CRC.
Thanks to the phase rotation in complex domain without decaying the amplitude, the real-valued output of the complex-valued RC shows periodic behavior inherently.
This natural property helps the proposed framework analyze the periodic pHRI.

For verification of the proposed framework, a rope-rotation/swinging experiment, in which a human and a direct-drive actuation system collaborate forcefully through a rope to rotate or swing it, is analyzed.
Its dataset contains eight types of motions: rotation or swinging; with slow or fast speed; and by left or right arm.
The proposed framework is trained with this dataset in order to generate the new but similar trajectories to ones in the dataset.
As a result, the proposed framework successfully constructs the eight clusters corresponding to the dataset's motions in the extracted latent state space, even though it does not receive any label information of the dataset's motions.
Such a well-distinguished latent state space enables the proposed framework to achieve the best prediction accuracy for the human observations and the robot actions, while the performance is deteriorated when either of the critical components in the proposed method is missing.

\section{Proposed framework}

\subsection{Overview}

\begin{figure}[tb]
    \centering
    \includegraphics[keepaspectratio=true,width=0.95\linewidth]{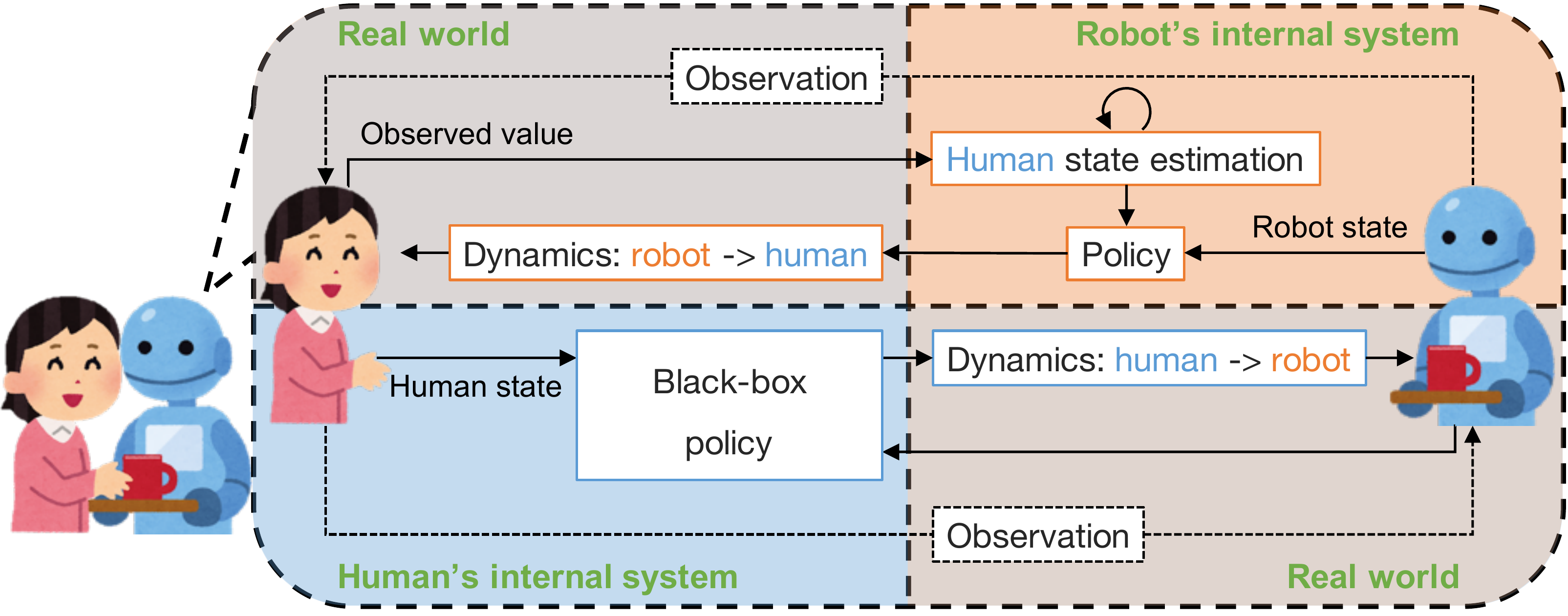}
    \caption{Conceptual framework for pHRI:
        in the robot's internal system, the human state is estimated based on the high-dimensional observation of the human;
        the robot then decides the optimal action for pHRI based on the estimated human state and own state, resulting in update of the human state;
        although the human's internal system is unknown, it would also output the action, which updates the robot state, according to the human and robot states;
        in this paper, only the upper parts of this framework, i.e. the modules related to pHRI from robot to human, are learned for the analysis of pHRI.
    }
    \label{fig:concept_phri}
\end{figure}

The pHRI model in our framework assumes to be capable of revealing human state from high-dimensional observation, although the method to reveal it is unknown.
Therefore, one should estimate the human state from the current observation and the history of interaction.
The human state transition dynamics cannot be predesigned as well, and should be learned.
Since the robot will take actions for human, the robot action policy should also be represented in relation to the human state estimation (and own state).
Such a conceptual framework is illustrated in Fig.~\ref{fig:concept_phri} with another direction of interaction (i.e. from human to robot).
Since this paper focuses on the analysis of pHRI, the proposed framework in this paper is built only for pHRI from robot to human.

Hence, to satisfy the above requirements, we consider the problem to learn the three modules for pHRI from robot to human simultaneously.
In particular, to easily capture the periodic features in pHRI, we contributes to design a new recurrent network (see the next section).
In this section, we derive a base of learning framework, i.e. a graphical model and its loss function to be minimized.

\subsection{Variational recurrent neural network}

To extract the latent space $z$ from the high-dimensional observation space $o$, variational autoencoder (VAE)~\cite{kingma2014auto,higgins2016beta} is frequently employed.
VAE derives a maximization problem of variational lower bound, and its variant, named $\beta$-VAE, regards it as a constrained optimization problem, which introduces the hyperparameter $\beta$ for regularization to prior information.
By integrating VAE with RNNs, such as long short-term memory (LSTM)~\cite{hochreiter1997long} and stochastic continuous-time RNN (S-CTRNN)~\cite{murata2013learning}, to handle time-series data (e.g. sampled from pHRI), variational recurrent neural network, so-called VRNN, has been developed~\cite{chung2015recurrent}.

Let us briefly introduce the optimization problem of VRNN.
At first, a sequence of observation up to $t$ time step, $o_{\leq t}$, is compressed to a historical feature $h_t$ by RNNs (originally, the standard RNN is employed~\cite{chung2015recurrent}).
\begin{align}
    o_{\leq t} \simeq h_t = \mathrm{RNN}(o_t, h_{t-1})
    \label{eq:rnn_obs}
\end{align}
Note that this formula ignores an argument $z_t$, which is added in the original paper, to simplify the computational graph for backpropagation and to improve realtimeness.

With $h_t$, we consider the problem of maximizing the prediction probability of $o_{t+1}$, i.e. $p(o_{t+1} \mid h_t)$.
Assuming that $o_{t+1}$ is generated according to a stochastic latent variable $z_{t+1}$ that is conditional on $h_t$, the variational lower bound is given as follows:
\begin{align}
    \ln p(o_{t+1} \mid h_t) &= \ln \int p(o_{t+1} \mid z_{t+1}) p(z_{t+1} \mid h_t) dz_{t+1}
    \nonumber \\
    &= \ln \int q(z_{t+1} \mid o_{t+1}, h_t) p(o_{t+1} \mid z_{t+1})
    \nonumber \\
    &\times \frac{p(z_{t+1} \mid h_t)}{q(z_{t+1} \mid o_{t+1}, h_t)} dz_{t+1}
    \nonumber \\
    &\geq \mathbb{E}_{q(z_{t+1} \mid o_{t+1}, h_t)}[\ln p(o_{t+1} \mid z_{t+1})]
    \nonumber \\
    &- \mathrm{KL}(q(z_{t+1} \mid o_{t+1}, h_t) \| p(z_{t+1} \mid h_t))
    \nonumber \\
    &= - \mathcal{L}_\mathrm{vrnn}
    \label{eq:loss_vrnn}
\end{align}
where $p(o_{t+1} \mid z_{t+1})$, $p(z_{t+1} \mid h_t)$, and $q(z_{t+1} \mid o_{t+1}, h_t)$ denote the decoder, time-dependent prior, and encoder, respectively.
Note that as well as eq.~\eqref{eq:rnn_obs}, the decoder is simplified by ignoring the direct dependency of $h_t$ for improving realtimeness.
The expectation operation in the first term is approximated by one-step Monte Calro method.
The second term, Kullback-Leibler (KL) divergence, is often analytically computed using the simple stochastic model, which has the closed-form solution of KL divergence, for the prior and encoder.

Specifically, the decoder and encoder are approximated by deep neural networks (DNNs) with the simple stochastic model like normal distribution.
When a new observation $o_{t+1}$ is obtained, $o_{t+1}$ and $h_t$ are fed into the networks and $\mathcal{L}_\mathrm{vrnn}$ is computed.
$o_{t+1}$ is also fed into RNNs to update $h_t$ to $h_{t+1}$ before a new observation arrives.
The total networks including RNNs are updated by minimization of $\mathcal{L}_\mathrm{vrnn}$ using a combination of backpropagation following the computational graph for computing $\mathcal{L}_\mathrm{vrnn}$ and one of the stochastic gradient descent (SGD) methods like~\cite{zaheer2018adaptive}.

\subsection{Extension for pHRI}

On pHRI, the observed information for human, $o$, is updated depending on not only the history of $o$ but also the robot's action, $a$.
In addition, the robot's action is generated (stochastically) by its feedback controller according to the current human state and/or feedforward controller according to the history of $a$.
Under such a natural problem statement, we extend VRNN to the one suitable for representing pHRI.

Specifically, the latent space $z$ is explicitly divided into the human state space $s$ and the robot action space $a$.
The decoder, encoder, and prior are also divided, and especially the encoders for the human state space and the robot action space are explicitly distinguished as the human state estimator and the robot policy (a.k.a. controller) $\pi$.
In addition, $s$ is assumed to be explicitly updated according to Markovian dynamics on pHRI.
\begin{align}
    s_{t+1} = f(s_t, a_t)
    \label{eq:dyn_gen}
\end{align}
where $f(\cdot)$ denotes the general nonlinear dynamics, which can be approximated by DNNs.

\begin{figure}[tb]
    \centering
    \includegraphics[keepaspectratio=true,width=0.95\linewidth]{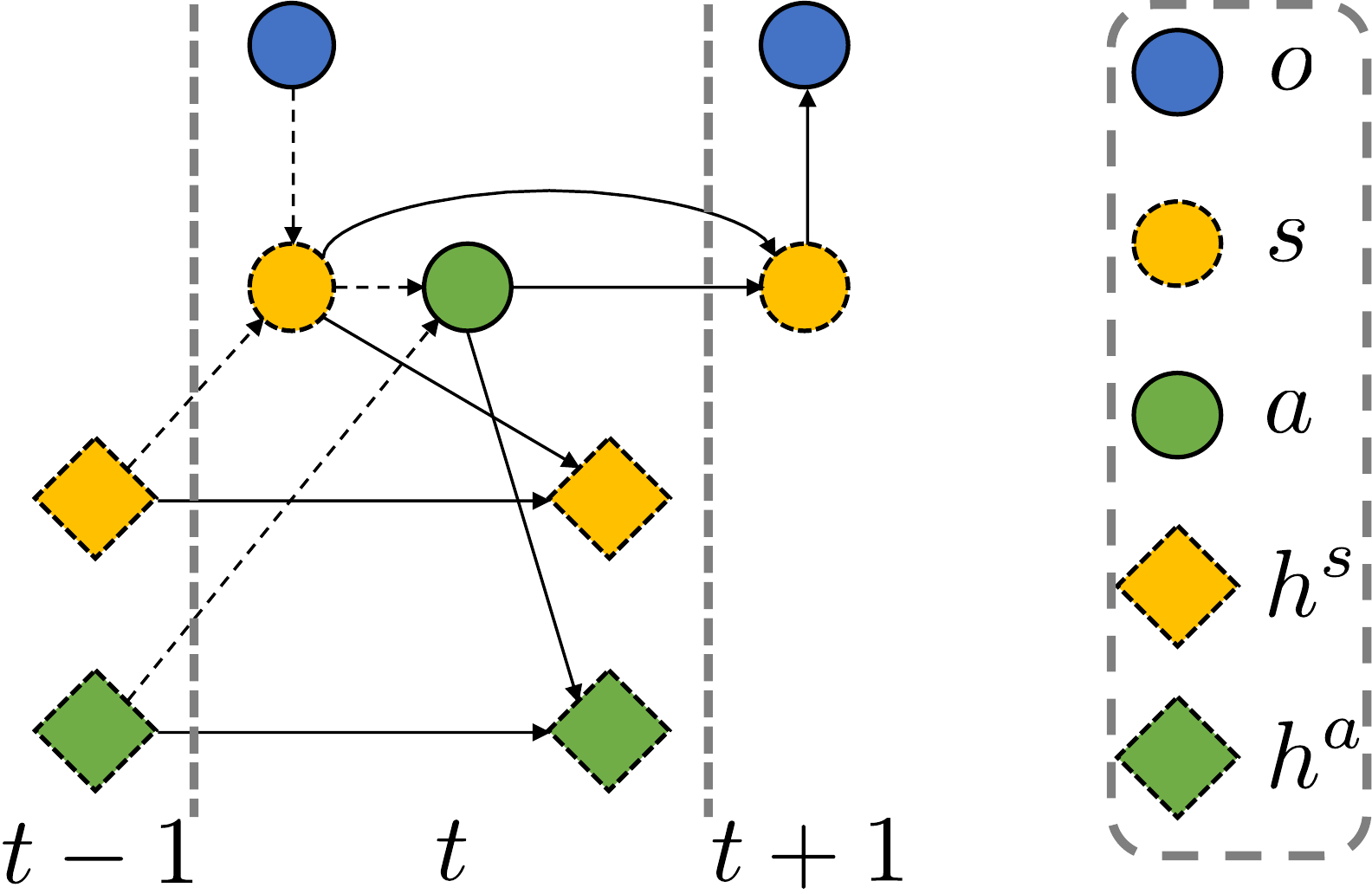}
    \caption{Graphical model of VRNN modified for pHRI:
        unobservable blocks have their frames drawn with dotted lines;
        dashed arrows indicate variational approximations;
        this model predicts future state and observation by relying on the current observation and information from the past;
        Markovian dynamics is explicitly assumed for handling pHRI.
    }
    \label{fig:img_vdyn}
\end{figure}

With these two conditions, eq.~\eqref{eq:loss_vrnn} is rederived as follows:
\begin{align}
    &\ln p(o_{t+1} \mid h_{t-1}^s, h_{t-1}^a)
    \nonumber \\
    &= \ln \iint p(o_{t+1} \mid s_{t+1})
    \nonumber \\
    &\times p(s_t \mid h_{t-1}^s) p(a_t \mid h_{t-1}^a) ds_t da_t
    \nonumber \\
    &= \ln \iint q(s_t \mid o_t, h_{t-1}^s) \pi(a_t \mid s_t, h_{t-1}^a)
    \nonumber \\
    &\times p(o_{t+1} \mid s_{t+1})
    \nonumber \\
    &\times \frac{p(s_t \mid h_{t-1}^s) p(a_t \mid h_{t-1}^a)}{q(s_t \mid o_t, h_{t-1}^s) \pi(a_t \mid s_t, h_{t-1}^a)} ds_t da_t
    \nonumber \\
    &\geq \mathbb{E}_{q(s_t \mid o_t, h_{t-1}^s) \pi(a_t \mid s_t, h_{t-1}^a)}[\ln p(o_{t+1} \mid s_{t+1})]
    \nonumber \\
    &- \mathrm{KL}(q(s_t \mid o_t, h_{t-1}^s) \| p(s_t \mid h_{t-1}^s))
    \nonumber \\
    &- \mathrm{KL}(\pi(a_t \mid s_t, h_{t-1}^a) \| p(a_t \mid h_{t-1}^a))
    \nonumber \\
    &= - \mathcal{L}_\mathrm{vdyn}
    \label{eq:loss_vdyn}
\end{align}
where $h_{t-1}^s$ and $h_{t-1}^a$ approximate the histories of $s$ and $a$ ($s_{\leq t-1}$ and $a_{\leq t-1}$) as the historical features, respectively, and can be updated by RNNs.
\begin{align}
    h_t^s &= \mathrm{RNN}(s_t, h_{t-1}^s)
    \label{eq:rnn_state} \\
    h_t^a &= \mathrm{RNN}(a_t, h_{t-1}^a)
    \label{eq:rnn_action}
\end{align}
The graphical model for this problem is illustrated in Fig.~\ref{fig:img_vdyn}.

\subsection{Auxiliary optimization for robot action space}

Although $s$ and $a$ are explicitly distinguished in the formulation of eq.~\eqref{eq:loss_vdyn}, when the total system is approximated by DNNs (and RNNs), $s$ and $a$ cannot be distinguished due to learning with backpropagation to the entire networks.
Therefore, an auxiliary optimization is provided for $a$, for which the supervised data can be collected, so that $s$ and $a$ have distinct roles.

To this end, the following optimization is additionally considered.
\begin{align}
    \mathcal{L}_\mathrm{act} &= \mathrm{KL}(\pi^*(a_t) \| \pi(a_t \mid s_t, h_{t-1}^a))
    \nonumber \\
    &\propto \mathbb{E}_{a_t \sim \pi^*}[- \ln \pi(a_t \mid s_t, h_{t-1}^a)]
    \label{eq:loss_act}
\end{align}
where $\pi^*$ denotes the true robot policy.
Since $\pi^*$ is at least a black-box function that can sample $a$, Monte Carlo approximation is available even if its KL divergence cannot be computed analytically.
In addition, by removing the term not related to the optimization of $\pi$, we actually obtain the expected value (i.e. sample mean) of the negative log-likelihood of $\pi$.

Following $\beta$-VAE~\cite{higgins2016beta}, the minimization problem for $\mathcal{L}_\mathrm{vdyn}$ in eq.~\eqref{eq:loss_vdyn} and $\mathcal{L}_\mathrm{act}$ in eq.~\eqref{eq:loss_act} can be regarded as the minimization problem constrained with the three KL divergences.
The weights $\beta_{1,2,3}$ are therefore introduced for the three KL divergences.
Consequently, given the trajectories of the tuples of the current observation, the robot's action, and the updated observation, $(o_t, a_t, o_{t+1})$, the loss function to be minimized is given as follows:
\begin{align}
    \mathcal{L} &= - \ln p(o_{t+1} \mid s_{t+1} = f(s_t, \tilde{a}_t))
    \nonumber \\
    &+ \beta_1 \mathrm{KL}(q(s_t \mid o_t, h_{t-1}^s) \| p(s_t \mid h_{t-1}^s))
    \nonumber \\
    &+ \beta_2 \mathrm{KL}(\pi(a_t \mid s_t, h_{t-1}^a) \| p(a_t \mid h_{t-1}^a))
    \nonumber \\
    &- \beta_3 \ln \pi(a_t \mid s_t, h_{t-1}^a)
    \label{eq:loss_total} \\
    \mathrm{s.t.}\ &
    \nonumber \\
    s_t &\sim q(s_t \mid o_t, h_{t-1}^s), \tilde{a}_t \sim \pi(a_t \mid s_t, h_{t-1}^a)
    \label{eq:dyn_gen2}
\end{align}
Note that $h_{t-1}^s$ and $h_{t-1}^a$ are updated using RNNs, as described in eqs~\eqref{eq:rnn_state} and~\eqref{eq:rnn_action}, at each time step.
In addition, $\tilde{a}_t$ is defined to distinguish it from $a_t \sim \pi^*$.

\section{Design of RNN suitable for periodic pHRI}

\subsection{Motivation}

\begin{figure}[tb]
    \centering
    \includegraphics[keepaspectratio=true,width=0.95\linewidth]{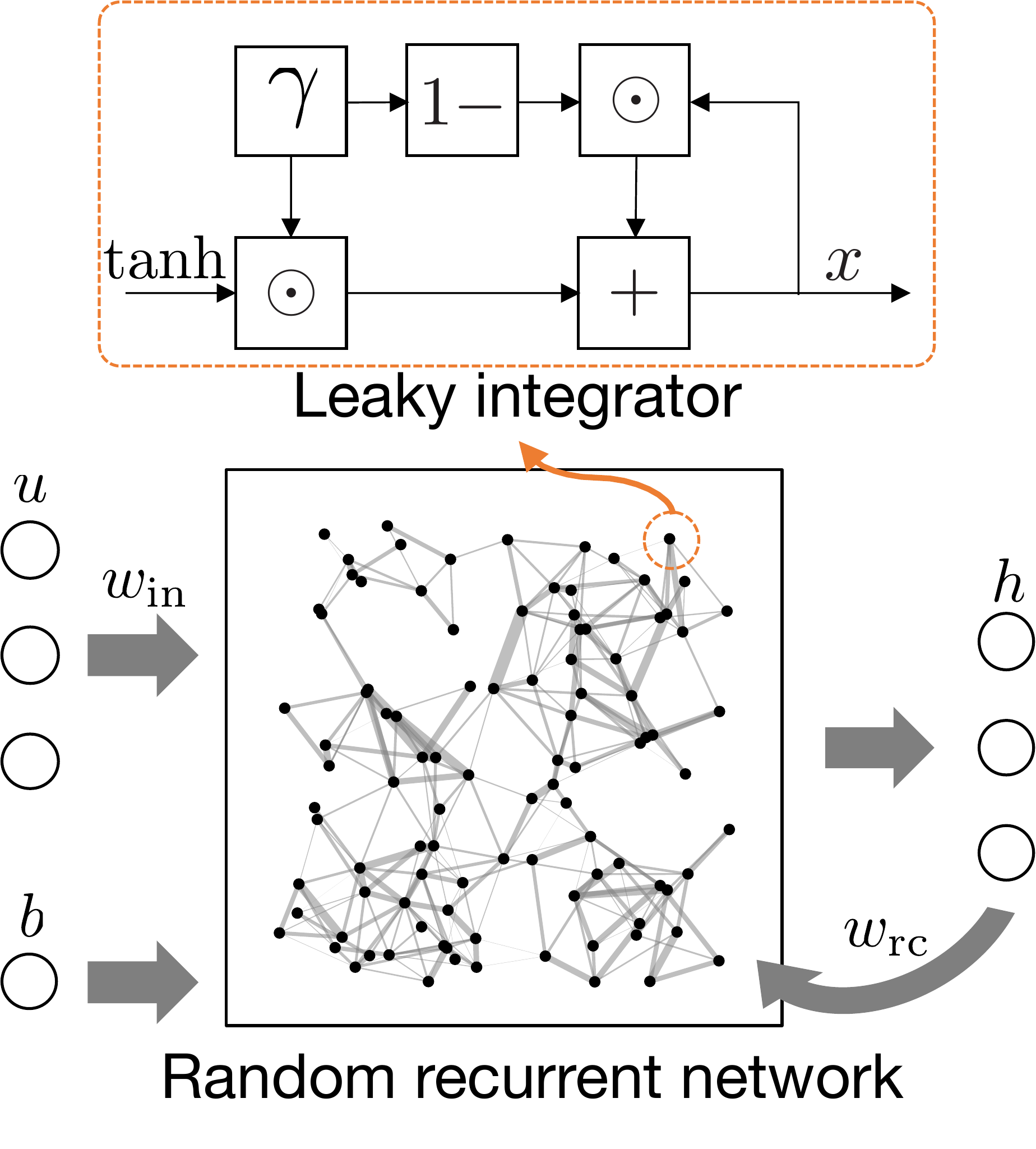}
    \caption{Basic structure of RC:
        RC has numerous neurons in a reservoir layer;
        each neuron connects with an input layer, a bias, and other neurons randomly;
        in addition, leaky integrator is applied for smooth update of its internal state;
        the internal states $h$ can be utilized as the time-series features.
    }
    \label{fig:img_rc}
\end{figure}

As described above, the proposed framework requires RNNs for $h_t^s$ and $h_t^a$.
The first bottleneck of the proposed framework is that, although the graphical model is simplified from the original~\cite{chung2015recurrent}, when backpropagation through time~\cite{werbos1990backpropagation} is applied for optimizing RNNs, a huge complex computational graph is constructed.
This increases the computational cost and induces learning instability especially in the problems, where RNNs cannot be optimized directly.
Although it is possible to keep the computational cost by using truncated BPTT~\cite{puskorius1994truncated}, it has been reported that the truncation causes a bias in the learning results~\cite{tallec2017unbiasing}.

In order to simplify the backpropagation and make the training of the system easier and more stable, we employ RC~\cite{jaeger2004harnessing,lukovsevivcius2009reservoir} as one of the RNNs (see Fig.~\ref{fig:img_rc}).
Although RC can provide sufficient time representation thanks to numerous random neurons, unlike LSTM~\cite{hochreiter1997long}, it is difficult to handle the prediction problems with long time delays.
For periodic pHRI, RC needs to be improved to capture the information of the longer-term past.
In this paper, we consider the extension of RC into complex domain.
The behavior in complex domain is represented by the decay of amplitude and the rotation of phase, and especially the rotation of phase leads to the utilization of the periodic past information.

\subsection{Complex-valued reservoir computing}

\begin{figure}[tb]
    \centering
    \includegraphics[keepaspectratio=true,width=0.95\linewidth]{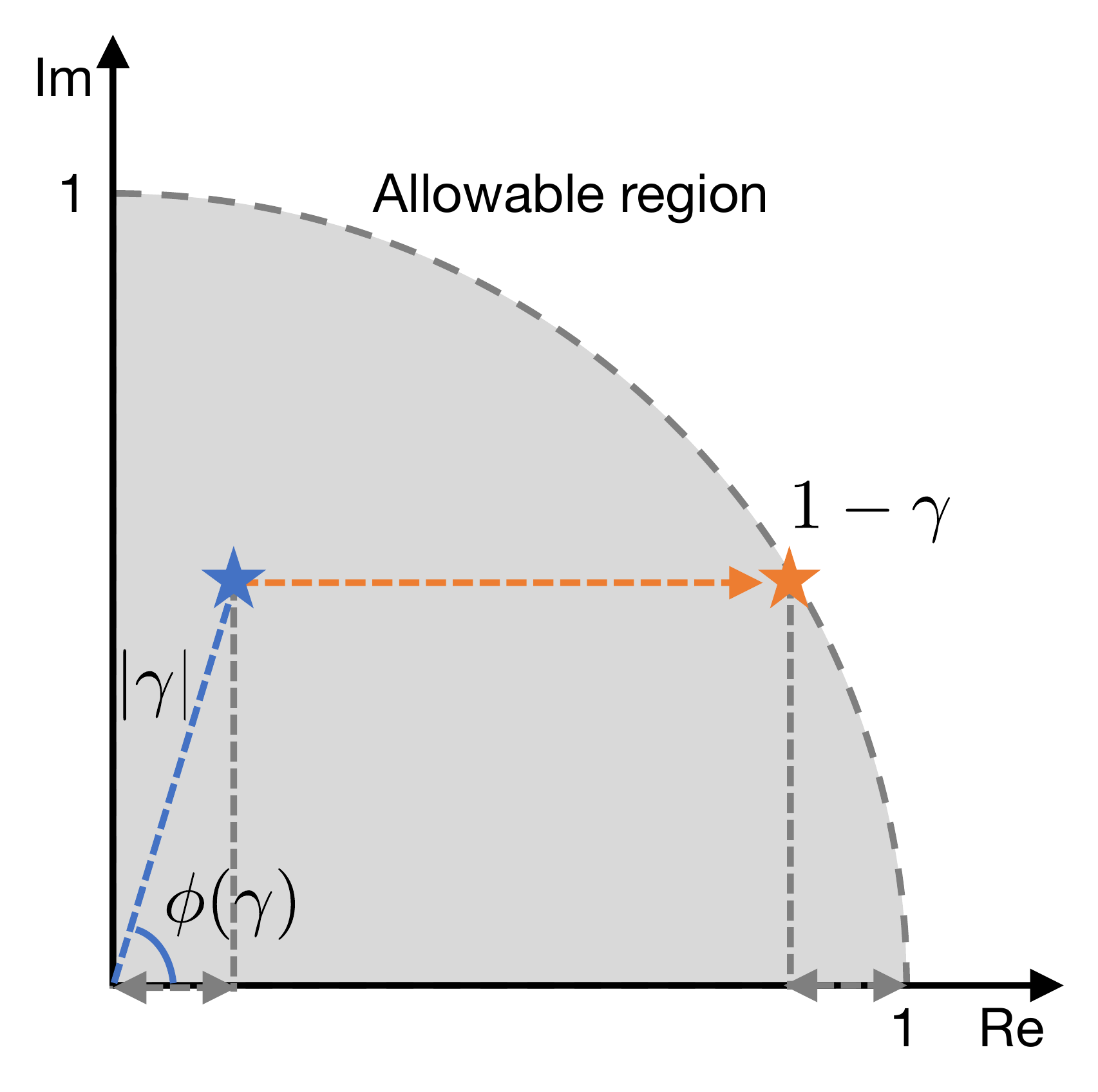}
    \caption{Geometric design of $\gamma$:
        for numerical stability, the amplitudes, $|\gamma|$ and $|1 - \gamma|$, must be less than $1$;
        if $|\gamma|$ is randomly given within $(0, 1]$ at first, the phase $\phi(\gamma)$ cannot be freely designed to satisfy $|1 - \gamma| < 1$;
        since we can easily find $1 - \gamma = 1 - |\gamma| \cos\phi(\gamma) + i |\gamma| \sin\phi(\gamma)$, the upper bound of $\phi(\gamma)$ can be solved as shown in eq.~\eqref{eq:phi_upper}.
    }
    \label{fig:img_cli}
\end{figure}

RC is a model built with fixed random weights, $w_\mathrm{in}$ and $w_\mathrm{rc}$, and biases, $b$, and updates its internal state $h$ according to input $u$ as follows:
\begin{align}
    h_{t+1} = (1 - \gamma) \odot h_t + \gamma \odot \tanh(w_\mathrm{in}^\top u_t + w_\mathrm{rc}^\top h_t + b)
    \label{eq:rc_gen}
\end{align}
where $\gamma \in (0, 1]$ represents the leaky integrator and $\odot$ denotes the element-wise multiplication.
Note that for numerical stability, the spectral radius of $w_\mathrm{rc}$ is desired to be less than one.

If RC has sufficient neurons and $w_\mathrm{in}$, $w_\mathrm{rc}$, $b$, and $\gamma$ are appropriately randomly given, a part of the neurons can adequately capture the features of the time-series data.
However, in principle, the RC defined above has an exponential decay of $h$, which makes it difficult to represent long-term characteristics (e.g. periodicity).
To overcome this drawback and make RC suitable for periodic pHRI, we focus on complex-valued neural networks~\cite{goh2007augmented,hu2012global}, which yield oscillatory behavior between real and imaginary values.
If RC is augmented into complex domain, such an oscillatory behavior gives the capability to represent periodic pHRI.

Specifically, $w_\mathrm{in}$, $w_\mathrm{rc}$, and $b$ can be given randomly as well even in complex domain, and since the spectral radius covers the complex matrix, numerical stability can be easily guaranteed~\cite{xia2008complex}.
As for the activation function, $\tanh$, the phase-amplitude version~\cite{hirose1992continuous} is employed.
\begin{align}
    \tanh(x) = \tanh(|x|)\exp(i \phi(x))
\end{align}
where $\phi(x)$ denotes the phase of $x$ in complex domain.

In contrast, $\gamma$ should be carefully designed since $1 - \gamma$ may violate $|1 - \gamma| < 1$ for numerical stability (see Fig.~\ref{fig:img_cli}).
To avoid this violation, the amplitude of $\gamma$, $|\gamma|$, is first randomly given.
Afterwards, the upper bound of the phase of $\gamma$, $\overline{\phi}(\gamma)$, is derived as follows:
\begin{align}
    (1 - |\gamma| \cos \overline{\phi}(\gamma))^2 + (|\gamma| \sin \overline{\phi}(\gamma))^2 = 1
    \nonumber \\
    \therefore \overline{\phi}(\gamma) = \arccos \cfrac{|\gamma|}{2}
    \label{eq:phi_upper}
\end{align}
The phase of $\gamma$, $\phi(\gamma)$, is therefore randomly given within $[0, \overline{\phi}(\gamma))$.

\subsection{Verification of qualitative behavior}

\begin{figure*}[tb]
    \centering
    \subfigure[Trajectories for all the neurons]{
        \includegraphics[keepaspectratio=true,width=0.31\linewidth]{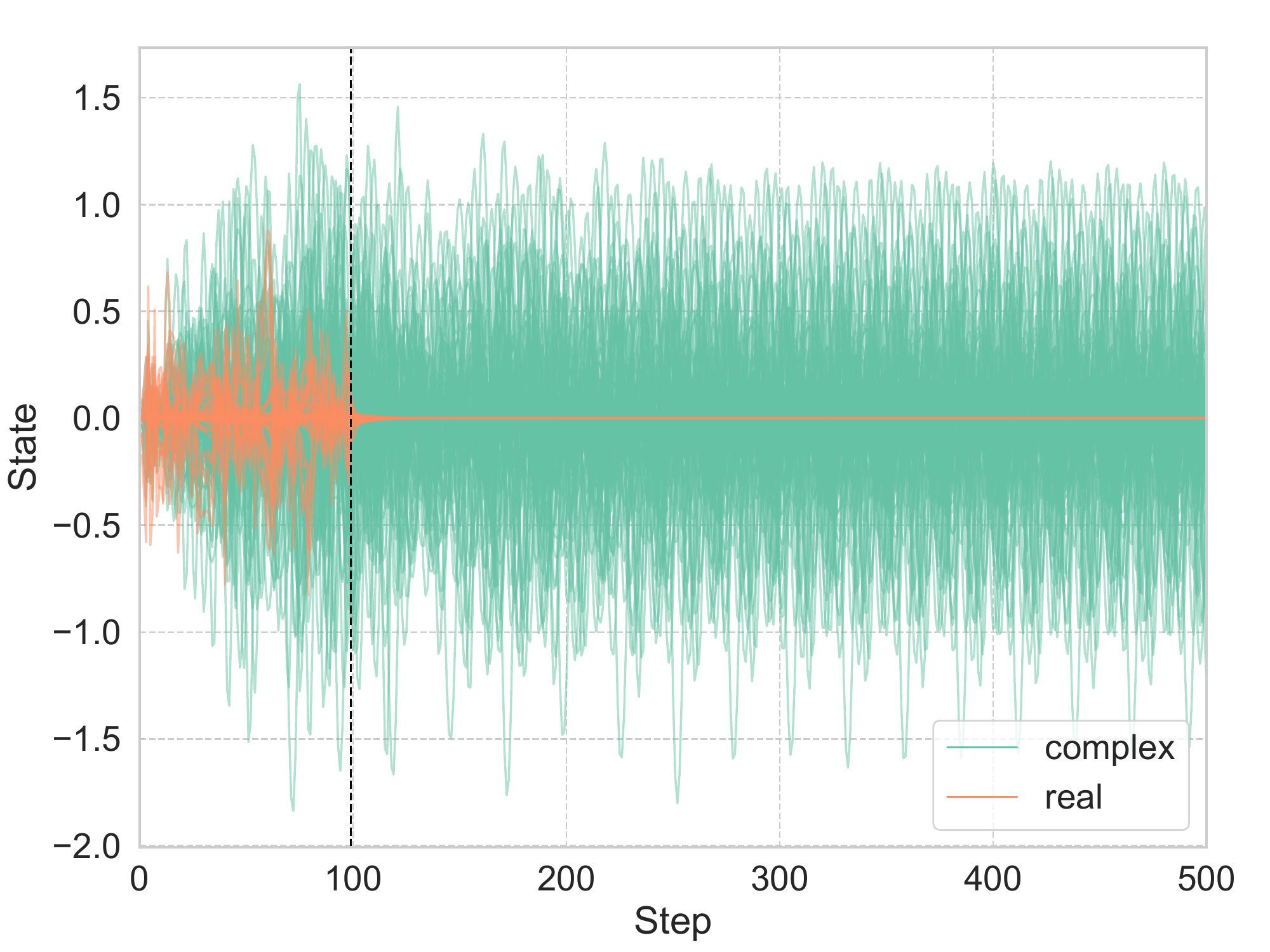}
    }
    \centering
    \subfigure[FFT analysis with random inputs]{
        \includegraphics[keepaspectratio=true,width=0.31\linewidth]{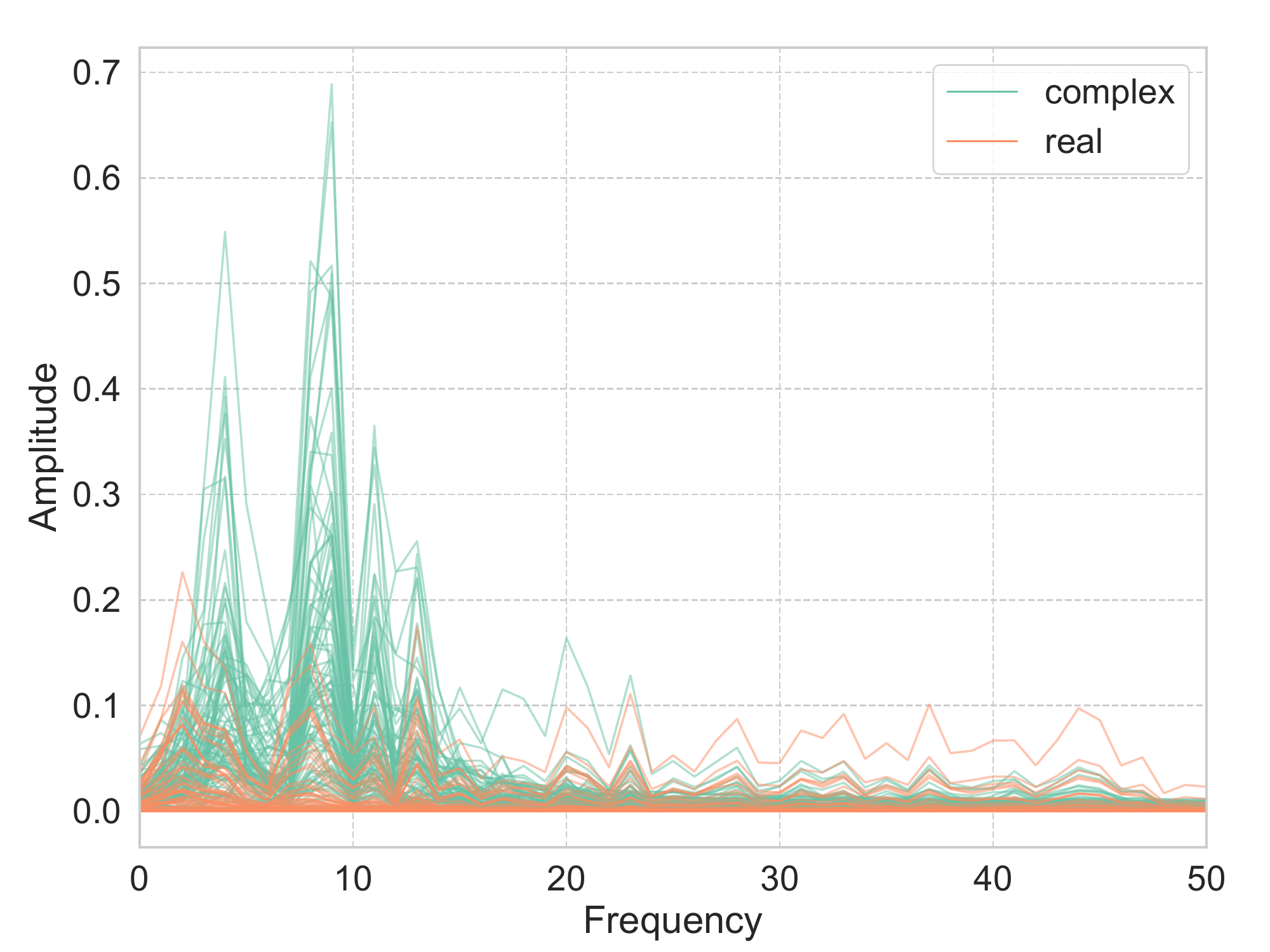}
    }
    \subfigure[FFT analysis of free response]{
        \includegraphics[keepaspectratio=true,width=0.31\linewidth]{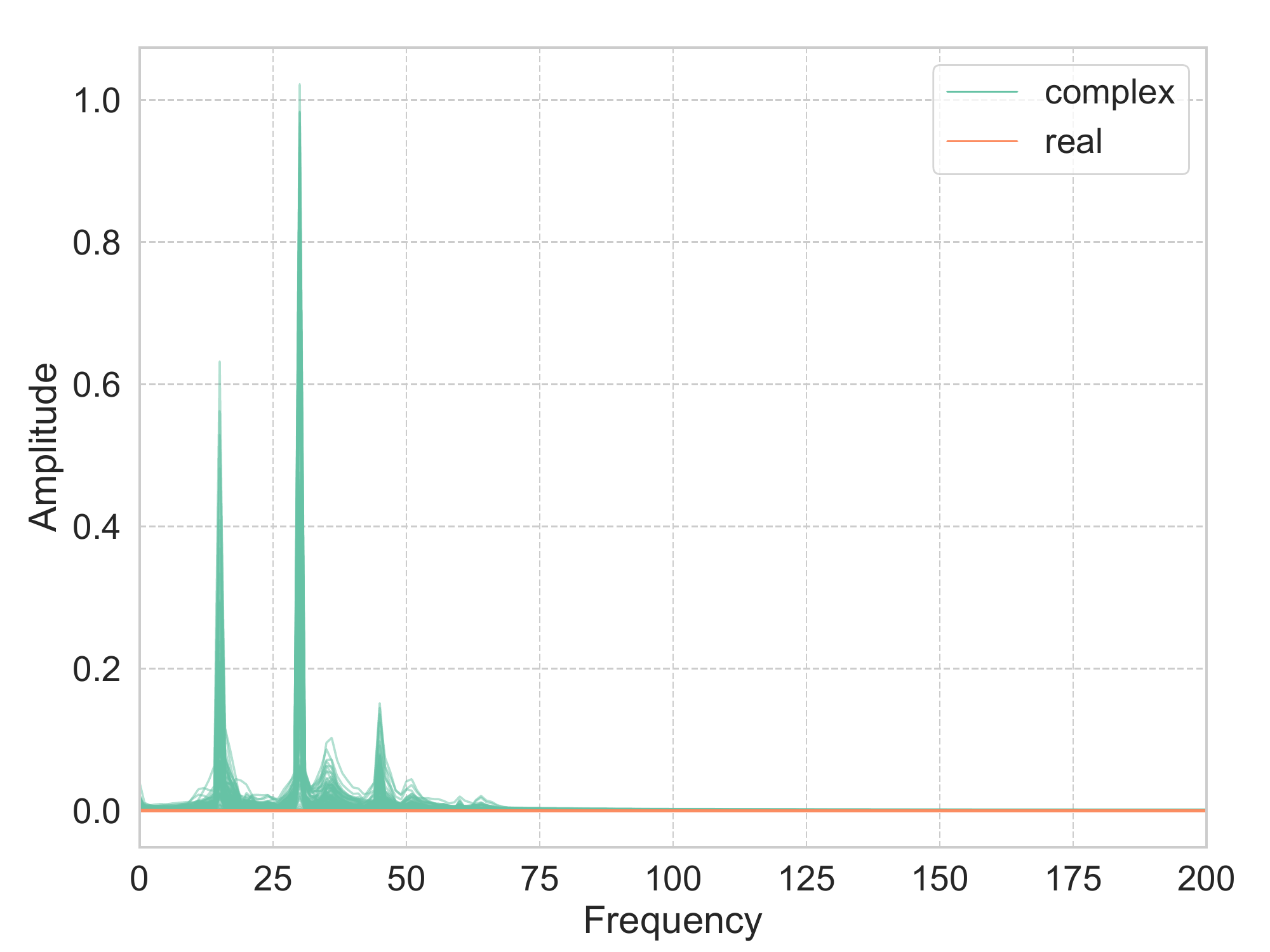}
    }
    \centering
    \caption{Example of oscillatory behavior on complex-valued RC:
        the bias term $b$ is set to zero in order to unify the convergence point of $h$ to zero for easier viewing;
        (a) although the standard RC excited only when the random inputs were applied until 100 steps, the complex-valued RC achieved the periodic oscillation without divergence during the free response;
        indeed, FFT analyses in (b) and (c) show that the complex-valued RC has the natural frequencies for the respective neurons, and amplifies $h$ with multiple frequencies.
    }
    \label{fig:sim_crc}
\end{figure*}

With the above implementation, we can see the oscillatory behavior in Fig.~\ref{fig:sim_crc}.
The free response after adding random inputs during the first 100 steps shows that the internal states continues to oscillate without decay.
In fact, by analyzing the frequency at this time through fast Fourier transform (FFT), multiple natural frequencies (probably, depending on the initialization) can be confirmed (see Fig.~\ref{fig:sim_crc}(c)).
Although the input does not zero during the actual pHRI like the first 100 steps, this internal property of the complex-valued RC can amplify the periodic signals as shown in Fig.~\ref{fig:sim_crc}(b) for the result of FFT during the first 100 steps with the random inputs.

As a remark, in the implementation, the complex-valued $h$ cannot be fed into the standard DNNs, hence, only the real value of $h$ is utilized for them, while the imaginary value of $h$ is still utilized for updating itself.
This restriction ensures that DNNs are always in real domain and can be updated using the usual learning rule without taking care of the fact that RC is in complex domain.

\section{Experiments}

\subsection{Rope rotation/swinging}

\begin{figure}[tb]
    \centering
    \includegraphics[keepaspectratio=true,width=0.95\linewidth]{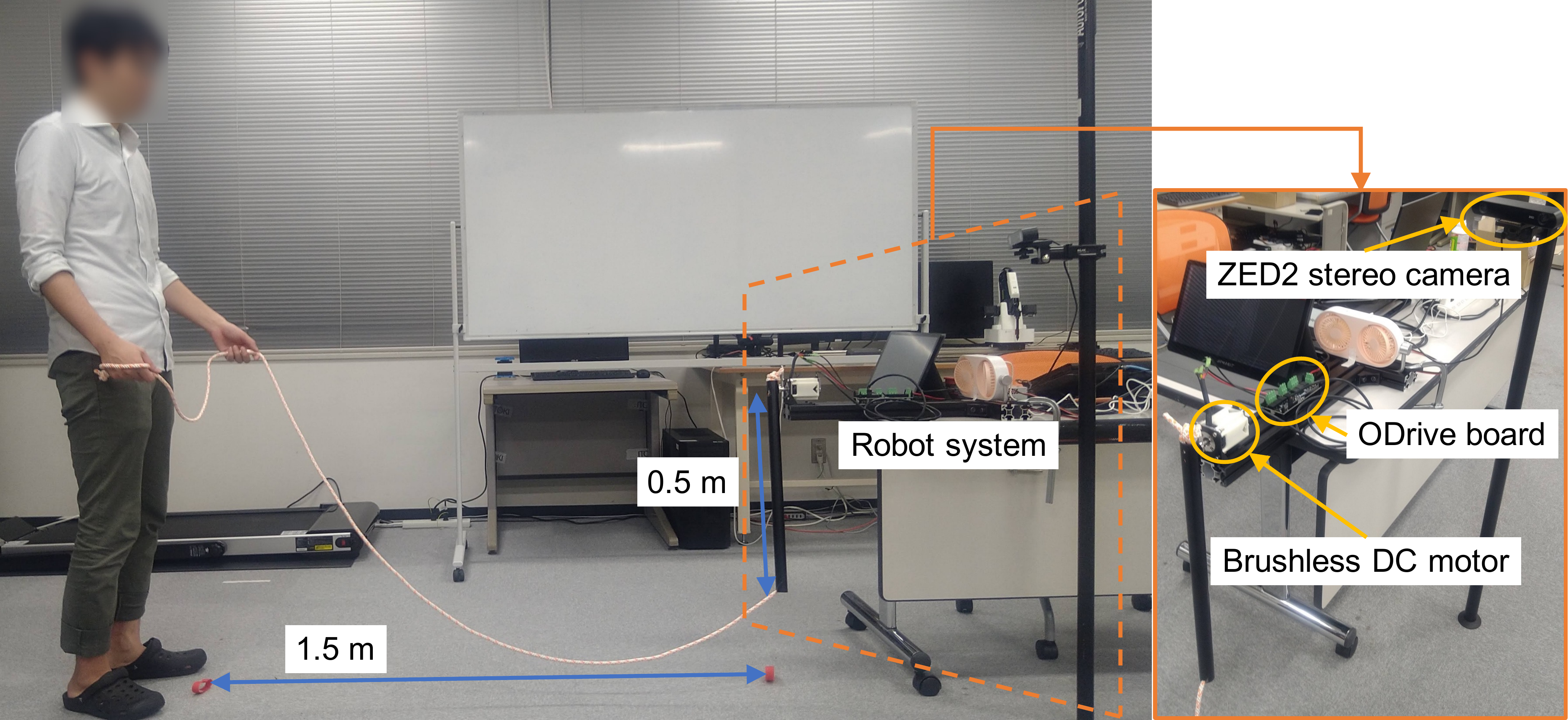}
    \caption{Experimental environment:
        a one-joint robot with a brushless DC motor attempts to rotate or swing a rope according to the specified conditions;
        A human partner attempts to follow its movements;
        the human motion can be detected by a ZED2 stereo camera.
    }
    \label{fig:exp_env}
\end{figure}

As an example of periodic pHRI, in this paper, we analyze a rope-rotation/swinging task.
The system configuration for this task is shown in Fig.~\ref{fig:exp_env}.

One side of the rope is attached to a one-joint robot with 0.5~m arm length and a brushless DC motor (D5065) controlled via ODrive board both of which are developed by ODrive Robotics.
Thanks to the absence of gears, this motor has back-drivability.
The robot accepts three types of commands: motor angle, angular velocity, and torque commands.
These are generated by feedforward rotational motion and admittance control (for details, see Appendix~\ref{app:motor}).

A human partner holds another end of the rope at a distance of around 1.5~m from the robot.
To increase a variation of the motions to be analyzed, he generates the motions by switching either the left or right arm.
When the rope was too long, the uncertainty of the movement was likely to be large, hence the rope length is adjusted not to touch the ground when grasped.

For human observation, we use a ZED2 stereo camera from Stereolabs.
It can detect the 3D coordinates of the 18 feature points of human and the center position and velocity of human; in total, 60 dimensions are acquired as the human observation space.
From them, the latent state space required for this task is extracted as three dimensions (for easy visual analysis).

Both the control and observation frequencies are set to be 30~fps (i.e. the time step $\Delta t \simeq 1/30$ sec), and each time-series trajectory has 900 time steps (i.e. around 30~sec).
In each trajectory, the robot operates in four different conditions, i.e. \textit{rotation} or \textit{swinging} with \textit{slow} or \textit{fast} speed (see Appendix~\ref{app:motion}), and the partner acts accordingly.
With switching the arm that mainly moves, there would be eight conditions in total.
We will verify whether the proposed framework can classify these conditions in the revealed latent state space.
Eleven trajectories on each condition (i.e. 88 trajectories with 9,900 data for all the conditions) are recorded for training, and three trajectories for validation and three trajectories for test are also done on each condition.

\subsection{Learning conditions}

\begin{figure}[tb]
    \centering
    \includegraphics[keepaspectratio=true,width=0.95\linewidth]{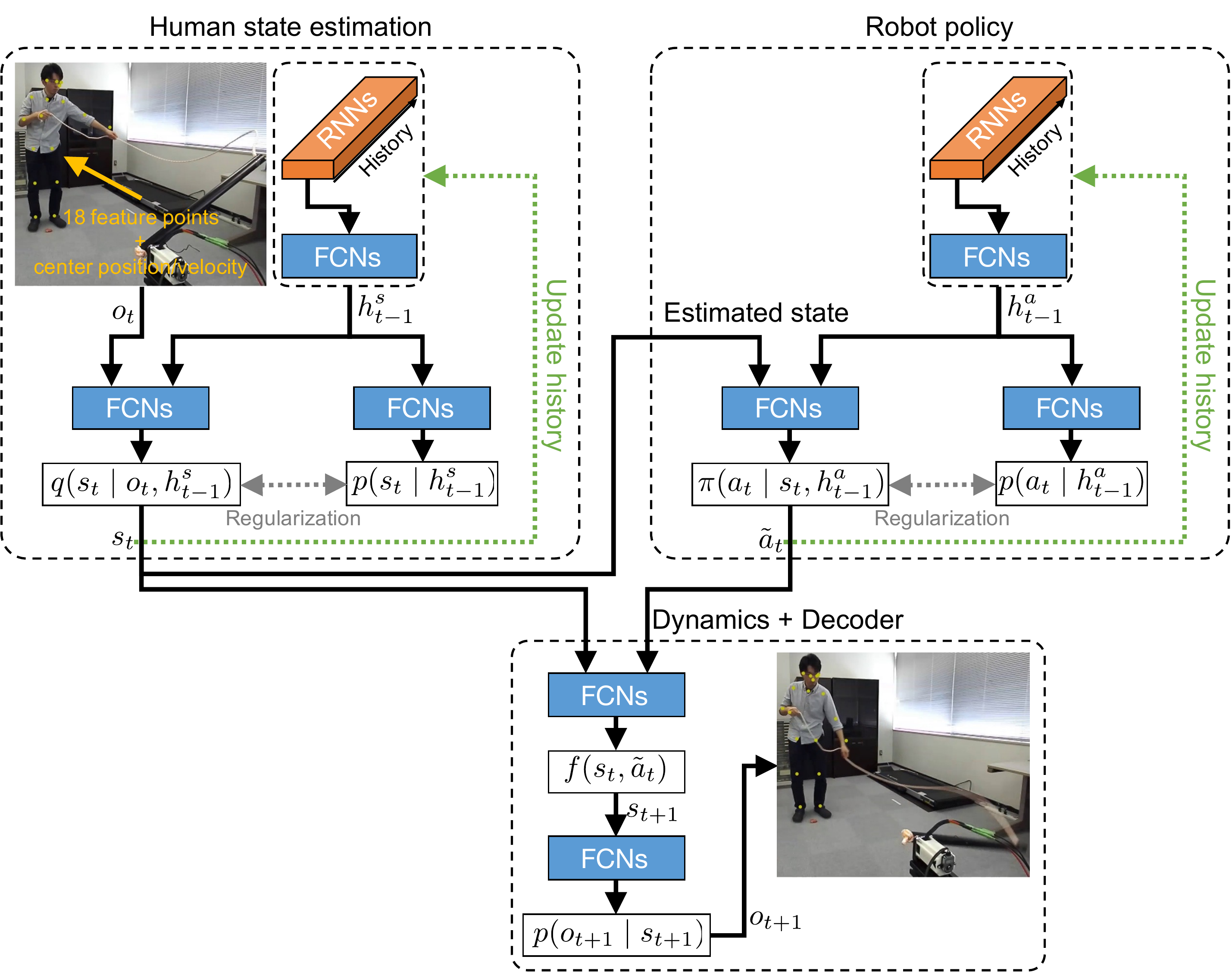}
    \caption{Neural network architecture:
        a human state estimation module outputs the estimated state according to the current observation and the past states;
        a robot policy module infers the exerted action in the current (estimated) state and the past actions;
        a dynamics module updates the state to the next one by acting the outputted action, then resulting in prediction of the new observation;
        we employ fully connected networks except RNN modules to map inputs to outputs;
        in addition, we apply nonlinear transformations to output parameters in the correct domain (e.g. a scale parameter in positive domain);
        note that the FCNs with two input arrows concatenate them as a vector signal.
    }
    \label{fig:network}
\end{figure}

To implement the proposed framework as a neural network architecture shown in Fig.~\ref{fig:network}, we utilize PyTorch~\cite{paszke2017automatic}.
For fully connected networks (FCNs) in the architecture, we design two layers with 100 neurons for each and Layer Normalization~\cite{ba2016layer} and ReLU function as a nonlinear activation function.
The decoder $p(o \mid s)$ is parameterized by diagonal student-t distribution for robust learning against observation noise and fluctuations in human behavior~\cite{takahashi2018student}, and the other distributions are parameterized by diagonal normal distribution.

For initializing (C)RCs in the proposed framework, each module has $N_\mathrm{rc} = 1000$ neurons.
All the parameters are sampled from uniform distribution $\in [-1, 1]$.
In addition, the parameters representing the recurrent network dynamics, $w_\mathrm{in}$, $w_\mathrm{rc}$, and $b$, have randomly sparse connections: each component of them is forced to be zero with probability $1 - 1/N_\mathrm{rc}^{0.9}$ based on~\cite{wood2012universality}.
Afterwards, we transform them so that all of them are within ranges to be satisfied for each variable.
That is, $w_\mathrm{rc}$ should make its spectral radius less than 1 (0.9 in this paper); $w_\mathrm{in}$ and $b$ are divided by $1 + \mathrm{rows}(w_\mathrm{in})$ ($\mathrm{rows}(\cdot)$ denotes the number of rows) to normalize them; $|\gamma|$ should be set to (0, 1]; and $\phi(\gamma)$ should be set to [0, $\overline{\phi}(\gamma)$) as obtained in eq.~\eqref{eq:phi_upper}.
Note that if the history information outputted from RC is directly concatenated with $o$ or $s$, the history may become dominant due to its large dimensionality.
The output from RC is therefore fed into FCNs before concatenation to adjust the number of dimensions and scale.

For training the networks, we employ td-AmsGrad, one of the state-of-the-art optimizers that is robust to noise and outliers~\cite{ilboudo2020robust,kobayashi2020towards}, with default parameters except learning rate $\alpha$.
We notice again that the parameters of (C)RC (i.e. $w_\mathrm{in}$, $w_\mathrm{rc}$, $b$, and $\gamma$) are not updated.
Hyperparamters for learning are summarized in Table~\ref{tab:param_model}.

\begin{table}[tb]
    \caption{Parameters for learning model}
    \label{tab:param_model}
    \centering
    \begin{tabular}{ccc}
        \hline\hline
        Symbol & Meaning & Value
        \\
        \hline
        $\alpha$ & Learning rate & 0.0001
        \\
        -- & Batch size & 44
        \\
        -- & Number of epoch & 100
        \\
        $\beta_1$ & Weight for state regularization & 0.1
        \\
        $\beta_2$ & Weight for policy regularization & 0.1
        \\
        $\beta_3$ & Weight for auxiliary policy optimization & 1.0
        \\
        \hline\hline
    \end{tabular}
\end{table}

\subsection{Learning performance}

\begin{figure}[tb]
    \centering
    \subfigure[Robot action]{
        \includegraphics[keepaspectratio=true,width=0.95\linewidth]{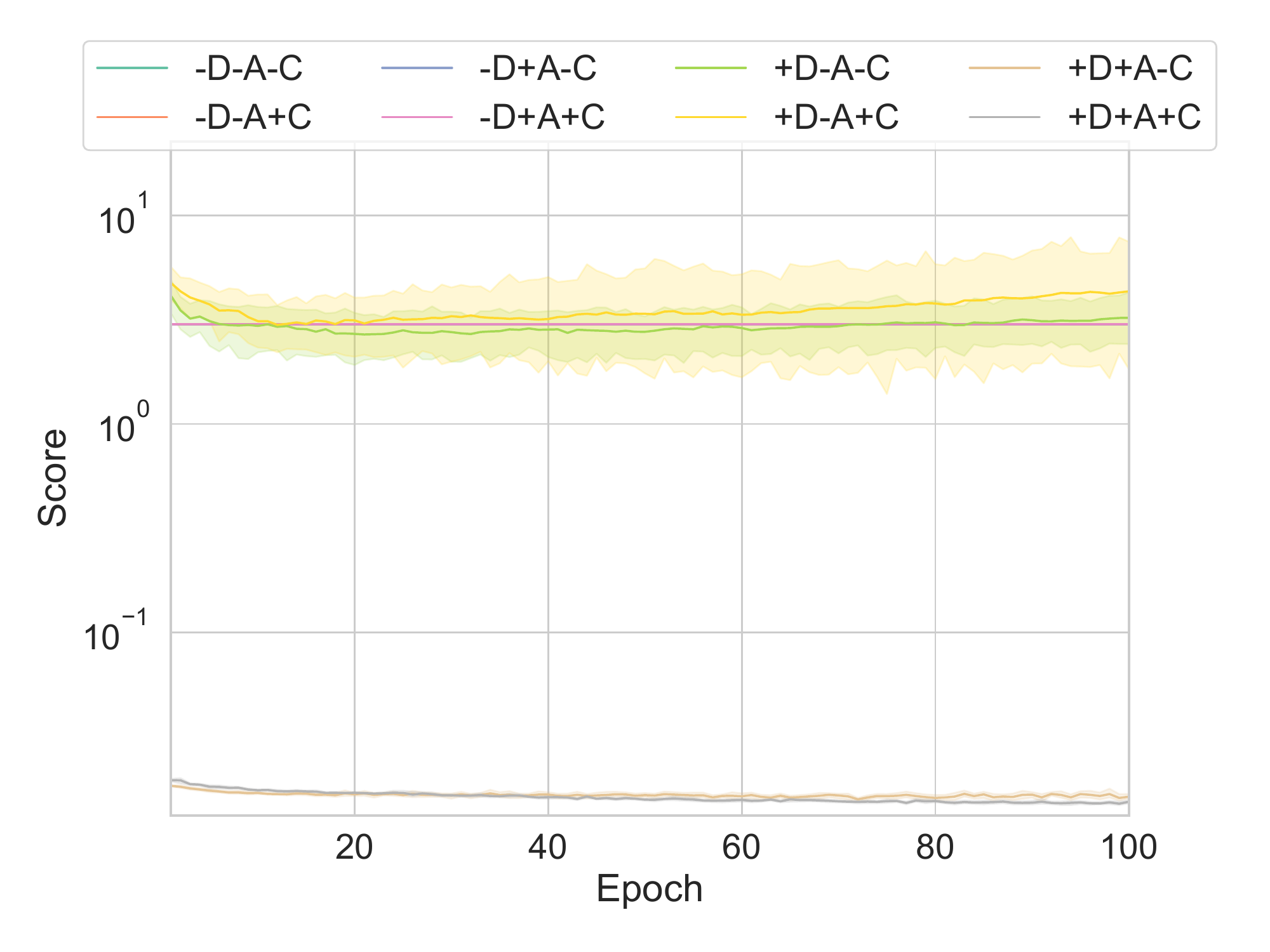}
    }
    \centering
    \subfigure[Human observation]{
        \includegraphics[keepaspectratio=true,width=0.95\linewidth]{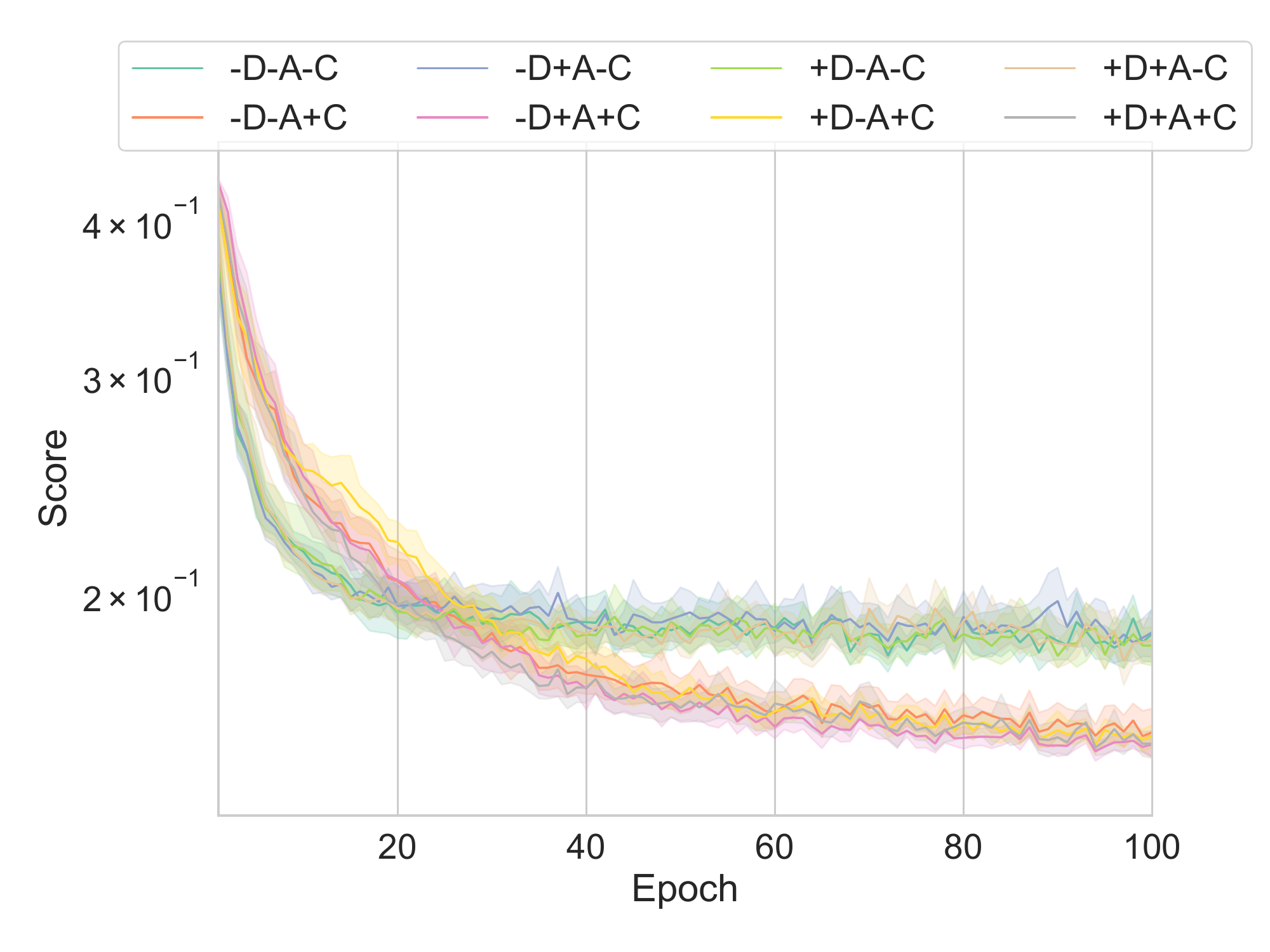}
    }
    \caption{Learnig curves of MSEs for the validation data:
        the vertical axes in both plots are with log scale;
        the labels `D', `A', and `C' indicate the respective components, and $\pm$ before them indicates the presence or absence of each component;
        from (a), we found that `D' and `A' must be required to predict the robot action;
        (b) indicated that the proposed CRC improved the prediction accuracy for the periodic motions.
    }
    \label{fig:exp_learn}
\end{figure}

\begin{table*}[tb]
    \caption{MSEs for the test data:
        the smaller is the better for all metrics;
        bold type indicates the best results.
    }
    \label{tab:exp_test}
    \centering
    \begin{tabular}{lccc|ccc}
        \hline\hline
         & & Robot action &  &  & Human observation &
        \\
        Condition & Mean & Median & STD & Mean & Median & STD
        \\
        \hline
        -D-A-C (conventional) & 3.012594 & 3.013845 & 0.005923 & 0.142771 & 0.132457 & 0.063805
        \\
        -D-A+C & 3.012594 & 3.013845 & 0.005923 & 0.118914 & 0.113092 & 0.042207
        \\
        -D+A-C & 3.012594 & 3.013845 & 0.005923 & 0.139802 & 0.132671 & 0.056632
        \\
        -D+A+C & 3.012594 & 3.013845 & 0.005923 & 0.116266 & 0.112286 & 0.041361
        \\
        +D-A-C & 3.232816 & 3.087314 & 1.910130 & 0.134759 & 0.129653 & 0.051112
        \\
        +D-A+C & 4.332716 & 1.514373 & 6.137522 & 0.117815 & 0.113291 & 0.041569
        \\
        +D+A-C & 0.016927 & 0.018676 & 0.008888 & 0.140475 & 0.130557 & 0.060881
        \\
        +D+A+C (proposed) & \textbf{0.015457} & \textbf{0.016703} & 0.008322 & \textbf{0.115742} & \textbf{0.110450} & 0.041228
        \\
        \hline\hline
    \end{tabular}
\end{table*}

The proposed framework consists of VRNN and three additional components:
`D' explicit human-robot dynamics, `A' auxiliary policy optimization, and `C' CRC for periodicity.
Note that, in the case with -D+A, $\tilde{a}_t$ is replaced to one vector with the same dimension to forcibly exclude the effects of $\tilde{a}_t$ on $s_{t+1}$.
We investigate the effects of them in terms of learning performance, i.e. mean squared error (MSE) between true and predicted human observations and/or robot actions.
The neural networks shown in Fig.~\ref{fig:network} are trained after being initialized with 20 random seeds for each of eight conditions (i.e. with and without the three components).
The performances for the validation data in the learning process are depicted in Fig.~\ref{fig:exp_learn}.
In addition, the performances for the test data are evaluated after training, and are summarized in Table~\ref{tab:exp_test}.

As can be seen in Fig.~\ref{fig:exp_learn}(a), the framework fails to learn the robot policy unless `D' and `A' are introduced.
As mentioned before, `D' alone does not work at all because the robot action $a$ is mixed into the human state $s$ without explicit discrimination of them.
On the other hand, if `A' is introduced, we expect that learning can be accomplished in a supervised learning manner.
However, it failed probably because the next robot action cannot be predicted from the robot past action sequences.
By `A' clarifying the roles of $a$ and $s$, and by `D' conveying the gradient information about the prediction of the human observation to the robot policy, the proposed framework makes the robot policy learnable.
The gradient information is also traced back to the $s$ fed into the robot policy, implying the benefits of $s$ reshaped to the one suitable for $a$.

From Fig.~\ref{fig:exp_learn}(b), we can say that the proposed CRC (i.e. ones with +C labels) outperformed the standard RC (i.e. ones with -C labels).
Although `C' caused higher MSE in the early stages of learning, that was reversed around 25 epoch.
Afterwards, the standard RC mostly remained on the local solution at that time; in contrast, the proposed CRC continued to reduce MSE even at 100 epoch (i.e. the end of learning).
The delay in learning was probably due to the more complex dynamics that is difficult to be disentangled.
The final result was due to the fact that parts of the complex dynamics were better suited to represent the periodic motions of this rope-rotation/swinging task, as we expected.

The same result for the test data can be seen from Table~\ref{tab:exp_test}.
The proposed framework with all components obtained the smallest metrics, mean and median of MSE.
This fact suggests that each component contributes not only to the aforementioned qualitative role for the robot policy (`D' and `A') or the human observation (`C'), but also to improving another prediction accuracy since the human and the robot interacted with each other under pHRI.

\subsection{Latent space analysis}

\begin{figure*}[tb]
    \centering
    \subfigure[-D-A-C (conventional)]{
        \includegraphics[keepaspectratio=true,width=0.95\linewidth]{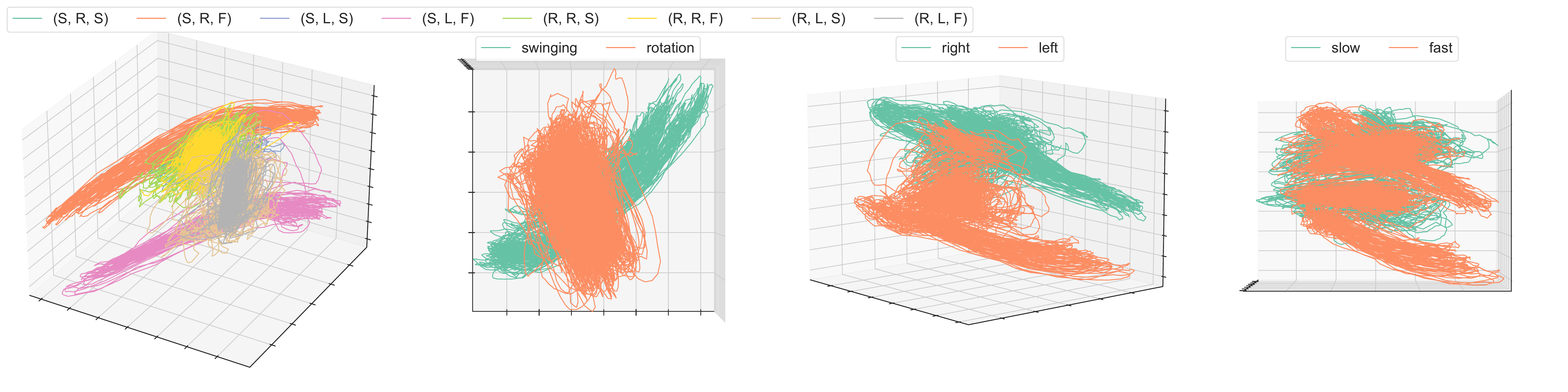}
    }
    \centering
    \subfigure[+D+A+C (proposed)]{
        \includegraphics[keepaspectratio=true,width=0.95\linewidth]{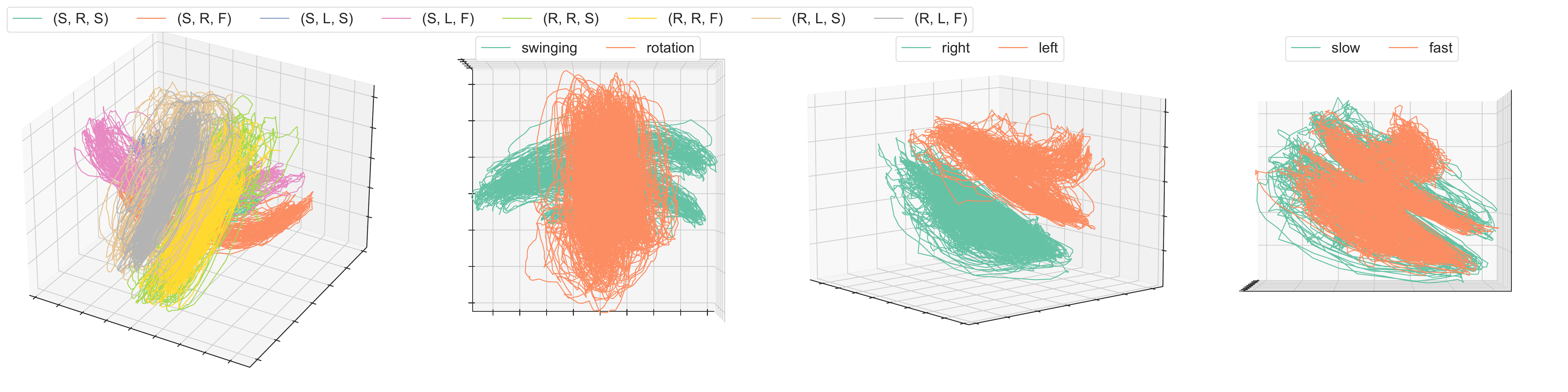}
    }
    \caption{Trajectories of test data in latent state space:
        tuples in the legend, which is for the leftmost plot, indicate (\{swinging, rotation\}, \{left, right\}, \{slow, fast\});
        remaining plots with their own legends are split into two groups for each condition;
        due to unsupervised learning, the scale of each axis is not meaningful, then the ticks for each axis were removed;
        instead, the topology of the clusters is worthy of attention, although the placement of clusters is hard to see from a single viewpoint;
        looking at the right three plots, the proposed framework seems to capture the characteristics of each condition and construct the clusters better than the conventional framework.
    }
    \label{fig:exp_state}
\end{figure*}

\begin{figure}[tb]
    \centering
    \subfigure[-D-A-C (conventional)]{
        \includegraphics[keepaspectratio=true,width=0.425\linewidth]{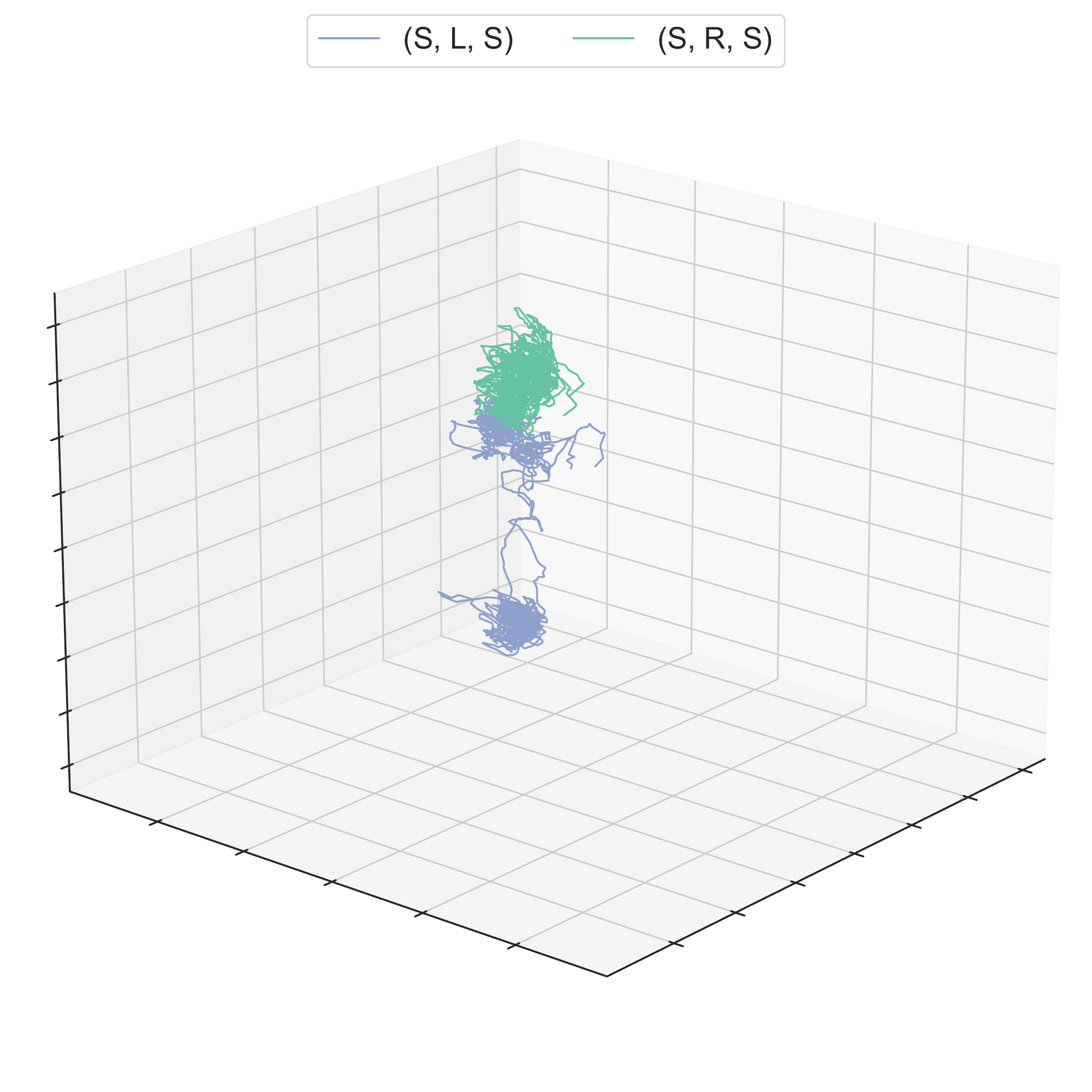}
    }
    \centering
    \subfigure[+D+A+C (proposed)]{
        \includegraphics[keepaspectratio=true,width=0.425\linewidth]{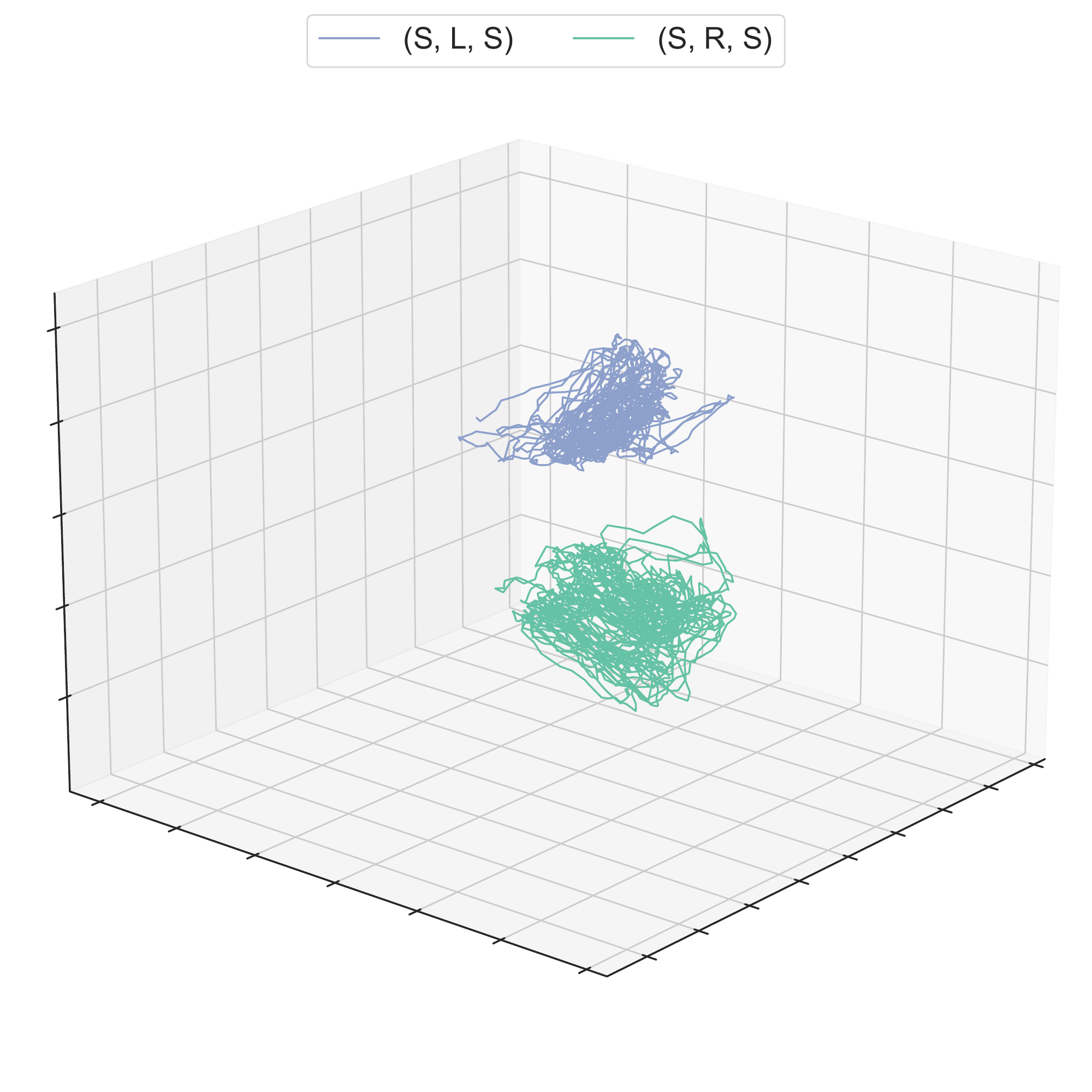}
    }
    \caption{Extracted trajectories of (S, L, S) and (S, R, S) from Fig.~\ref{fig:exp_state}:
        (a) the conventional method failed to make a cluster of (S, L, S), namely, one of the trajectories was placed on the cluster of (S, R, S);
        (b) the proposed method clearly succeeded in distinguishing the two clusters.
    }
    \label{fig:exp_diff}
\end{figure}

Finally, for the proposed framework with all components, we analyze its latent state space.
Since it is difficult to include all the data due to the limitation of space, the three trajectories for each condition in the test data are converted to $s$ using the model initialized with the second random seed, which resulted in the similar accuracy to the mean and median in Table~\ref{tab:exp_test}.
The transformed trajectories in the latent state space is illustrated in Fig.~\ref{fig:exp_state}.
Note that the -D-A-C model initialized with the second random seed, the accuracy of which was also close to the statistical metrics in Table~\ref{tab:exp_test}, was also used for transformation as the conventional method.
In addition, it was difficult to analyze the eight clusters in a 3D plot because they mostly overlapped from any angles, hence, for the sake of visibility, all the trajectories were split into two groups for each condition (e.g. swinging or rotation).

Although the role and scale of each axis are different in the two methods (even for each random seed) due to unsupervised learning, the swinging and rotation trajectories were arranged almost orthogonally.
While these two motions should have similar scenes if their moments are cut out, it is thought that such an orthogonal state space was obtained by appropriately considering the time series.

The trajectories driven mainly by left and right arms were symmetrically separated with the proposed framework.
However, with the conventional framework, some of the trajectories driven mainly by the left arm seem to be mixed up in the cluster for the right arm.
To confirm this, only the (S, L, S) and (S, R, S) trajectories were extracted and drawn on Fig.~\ref{fig:exp_diff}.
Indeed, the two clusters were clearly separated by the proposed framework, but in the conventional framework, one of the trajectories of (S, L, S) was misclassified into the cluster on (S, R, S).
This result suggests that the difference in the ability to represent the time-series data led to this result and also affected the prediction accuracy.

Finally, we focus on the two groups split according to the speed (i.e. slow or fast).
Compared to the other conditions, these two were not distinguished well, but the cluster for the fast motions was in the cluster for the slow motions.
To clarify that, an attached video shows the motions corresponding to the test data and the animations of the trajectories.
From the video, we found that, when rotating the rope, the cluster for the fast motions was certainly in the cluster for the slow motions.
In addition, when swinging the rope, the cluster for the slow motions was slightly shifted from the one for the fast motions while significantly reducing its amplitude.
The inclusion relation between the clusters for the rotation motions may be attributed to the fact that their basic motions were the same, but in the fast motion, the motion was compacted by actively using the inertia of the rope.

\section{Conclusion}

This paper developed a new data-driven framework for analyzing periodic pHRI in latent state space using the modified VRNN derivation and CRC.
Specifically, we modified VRNN in order to explicitly include the latent state dynamics updated according to the robot's action, which was also inferred from data.
As the RNN model suitable for representing periodic motions, we augmented the standard RC in real domain to the one in complex domain, named CRC.
The phase rotation in complex domain intuitively enables CRC to represent the periodic time-series data.
For verification of the proposed framework, a rope-rotation/swinging experiment was analyzed.
As a result, the highest observation and action prediction accuracy was obtained with the proposed method that included all three important components.
In addition, the proposed framework achieved the well-distinguished latent state space for analyzing the eight-type periodic motions, although the conventional method confused one of them.

Since the main focus of this paper was only on human motion analysis and classification, the robot policy was pre-designed (and imitated in the framework).
However, in real-world applications, the robot policy may be desired to be optimized toward the ideal human state.
To this end, in the future work, the inference of the human internal modules (see the lower of Fig.~\ref{fig:concept_phri}) is also necessary to predict overall pHRI, and by acquiring them, we expect the robot to be able to optimize its policy through model predictive control.

\appendix

\subsection{Motor control}
\label{app:motor}

The dynamics defined by the virtual impedance is employed to control the motor.
Specifically, an inertia $m$, a damping coefficient $c$, and a spring coefficient $k$ defines for the following dynamics.
\begin{align}
    &m \ddot{\theta} + c (\dot{\theta} - \dot{\theta}^\mathrm{ref}) + k (\theta - \theta^\mathrm{ref}) = \tau
    \nonumber \\
    &\therefore \ddot{\theta} = \frac{1}{m} \left \{ \tau - c (\dot{\theta} - \dot{\theta}^\mathrm{ref}) - k (\theta - \theta^\mathrm{ref}) \right \}
\end{align}
where the actual angle $\theta$ and the angular velocity $\dot{\theta}$ can be measured using an encoder, and the external torque $\tau$ is estimated as below.
$\dot{\theta}^\mathrm{ref}$ and $\theta^\mathrm{ref}$ are given to track the reference trajectory (see Appendix~\ref{app:motion}).

Since we use the brushless DC motor directly to control the robot, its measured current $i$ is in proportion to the exerted torque.
However, since the internal torque generated by the controller cannot be isolated from it, the system simply assumes that the increase or decrease from the time average of the exerted torque $\bar{\tau}$ corresponds to $\tau$.
\begin{align}
    \tau_i &= \kappa i
    \\
    \bar{\tau} &\gets \eta \bar{\tau} + (1 - \eta) \tau_i
    \\
    \tau &= \tau_i - \bar{\tau}
\end{align}
where $\kappa$ denotes the hardware torque constant and $\eta$ is designed for exponential moving average.

By Euler integration of the derived acceleration $\ddot{\theta}$, the command angular velocity and angle are obtained as follows:
\begin{align}
    \dot{\theta}^\mathrm{cmd} &= \dot{\theta} + \ddot{\theta} \Delta t
    \\
    \theta^\mathrm{cmd} &= \theta + \dot{\theta}^\mathrm{cmd} \Delta t
\end{align}
In order to weaken the resistance to $\tau$, the following torque command is heuristically given to the controller.
\begin{align}
    \tau^\mathrm{cmd} = - \mu \tau
\end{align}
where $\mu$ denotes the gain.
Note that since these command values differ in scale and are unsuitable for learning, the robot's action space is defined as the differences of them from the previous command values.

The parameters associated with the above controller design are listed up in Table~\ref{tab:param_motor}.

\begin{table}[tb]
    \caption{Parameters for motor control}
    \label{tab:param_motor}
    \centering
    \begin{tabular}{ccc}
        \hline\hline
        Symbol & Meaning & Value
        \\
        \hline
        $m$ & Inertia & 1.0
        \\
        $c$ & Damping coefficient & 50.0
        \\
        $k$ & Spring coefficient & 25.0
        \\
        $\eta$ & Smoothing factor & 0.9
        \\
        $\mu$ & Gain for anti-resistance & 0.5
        \\
        \hline\hline
    \end{tabular}
\end{table}

\subsection{Motion generation}
\label{app:motion}

To generate the reference trajectory for the rope rotation/swinging, we design two simple velocity profiles as below.
By Euler integration of the designed velocity $\dot{\theta}^\mathrm{ref}$, the reference angle $\theta^\mathrm{ref}$ is derived as follows:
\begin{align}
    \theta^\mathrm{ref} \gets \theta^\mathrm{ref} + \dot{\theta}^\mathrm{ref} \Delta t
\end{align}
Note that the motor we used counts the angle over one turn without resetting.

For the rope rotation, a trapezoidal velocity profile is employed.
Given an acceleration time $t_a$, a constant velocity time $t_c$, and a maximum velocity $v$, the following $\dot{\theta}^\mathrm{ref}$ at the current time $t$ is given.
\begin{align}
    \dot{\theta}^\mathrm{ref} = \begin{cases}
        v \frac{t}{t_a} & 0 \leq t < t_a
        \\
        v & t_a \leq t \leq t_a + t_c
        \\
        v - v \frac{t - (t_a + t_c)}{t_a} & t_a + t_c < t < 3 t_a + t_c
        \\
        - v & 3 t_a + t_c \leq t \leq 3 t_a + 2 t_c
        \\
        - v + v \frac{t - (3 t_a + 2 t_c)}{t_a} & 3 t_a + 2 t_c < t \leq 4 t_a + 2 t_c
    \end{cases}
\end{align}
In each trial, this profile is repeated by resetting $t$ until the trial ends.

For the rope swinging, a cosine velocity profile is employed.
Given a swing period $t_s$ and an amplitude of swinging (i.e. the maximum velocity $v$), the velocity profile is given as follows:
\begin{align}
    \dot{\theta}^\mathrm{ref} = v \cos \left (2\pi \frac{t}{t_s} \right )
\end{align}

The parameters associated with the above velocity profiles are listed up in Table~\ref{tab:param_motion}.

\begin{table}[tb]
    \caption{Parameters for velocity profiles:
        the two values in curly brackets are those used in the slow and fast conditions, respectively.
    }
    \label{tab:param_motion}
    \centering
    \begin{tabular}{ccc}
        \hline\hline
        Symbol & Meaning & Value
        \\
        \hline
        $t_a$ & Acceleration time & 1.0
        \\
        $t_c$ & Constant velocity time & 4.0
        \\
        $v$ & Maximum velocity for rotation & \{0.8, 1.2\}
        \\
        \hline
        $t_s$ & Swing period & 0.6$\pi$
        \\
        $v$ & Maximum velocity for swinging & \{0.5, 1.0\}
        \\
        \hline\hline
    \end{tabular}
\end{table}

\section*{ACKNOWLEDGMENT}

The research was supported by ROIS NII Open Collaborative Research 2020 (20S0701) and The Support Center for Advanced Telecommunications Technology Research Foundation (SCAT) Research Grant.

\bibliographystyle{IEEEtran}
{
\bibliography{biblio}
}

\end{document}